\documentclass[]{fairmeta}

\usepackage{amsmath}
\usepackage{amssymb}
\usepackage{mathtools}
\usepackage{amsthm}
\usepackage{algorithm}
\usepackage{algorithmic}
\usepackage{booktabs}
\usepackage{multirow}
\usepackage{makecell}
\usepackage[table]{xcolor}
\usepackage{tabularx}
\usepackage{array}
\definecolor{SparseRow}{HTML}{FFF6E5} 
\definecolor{BestTxt}{HTML}{1F4E79}   
\newcolumntype{L}{>{\raggedright\arraybackslash}X}
\usepackage{siunitx}

\usepackage{tikz}
\DeclareRobustCommand{\circnum}[1]{%
  \tikz[baseline=(C.base)]{%
    \node[draw,circle,inner sep=0.6pt,line width=0.4pt] (C) {\scriptsize #1};
  }%
}
\usepackage{adjustbox}

\title{Sparse Forcing: Native Trainable Sparse Attention for Real-time Autoregressive Diffusion Video Generation}

\author[1,2,*]{Boxun Xu}
\author[1]{Yuming Du}
\author[1]{Zichang Liu}
\author[1]{Siyu Yang}
\author[1]{Ziyang Jiang}
\author[1]{Siqi Yan}
\author[1]{Rajasi Saha}
\author[1]{Albert Pumarola}
\author[1]{Wenchen Wang}
\author[2]{Peng Li}

\affiliation[1]{Meta Superintelligence Labs}
\affiliation[2]{University of California, Santa Barbara}

\contribution[*]{Work done at Meta}

\abstract{We introduce Sparse Forcing, a training-and-inference paradigm for autoregressive video diffusion models that improves long-horizon generation quality while reducing decoding latency. 
Sparse Forcing is motivated by an empirical observation in autoregressive diffusion rollouts: attention concentrates on a persistent subset of salient visual blocks, forming an implicit spatiotemporal memory in the KV cache, and exhibits a locally structured block-sparse pattern within sliding windows.
Building on this observation, we propose a trainable native sparsity mechanism that learns to compress, preserve, and update these persistent blocks while restricting computation within each local window to a dynamically selected local neighborhood.
To make the approach practical at scale for both training and inference, we further propose Persistent Block-Sparse Attention (PBSA), an efficient GPU kernel that accelerates sparse attention and memory updates for low-latency, memory-efficient decoding. 
Experiments show that Sparse Forcing improves the VBench score by +0.26 over Self-Forcing on 5-second text-to-video generation while delivering a 1.11--1.17$\times$ decoding speedup and 42\% lower peak KV-cache footprint. 
The gains are more pronounced on longer-horizon rollouts, delivering improved visual quality with +0.68 and +2.74 VBench improvements, and 1.22$\times$ and 1.27$\times$ speedups on 20-second and 1-minute generations, respectively.}

\date{January 20, 2026}
\correspondence{Boxun Xu(\email{boxunxu@ucsb.edu})}
\metadata[Project Page]{\url{https://boxunxu.top/SparseForcing}}

\begin{document}

\maketitle

\section{Introduction}\label{sec:introduction}

\begin{figure*}
    \centering
    \includegraphics[width=\linewidth]{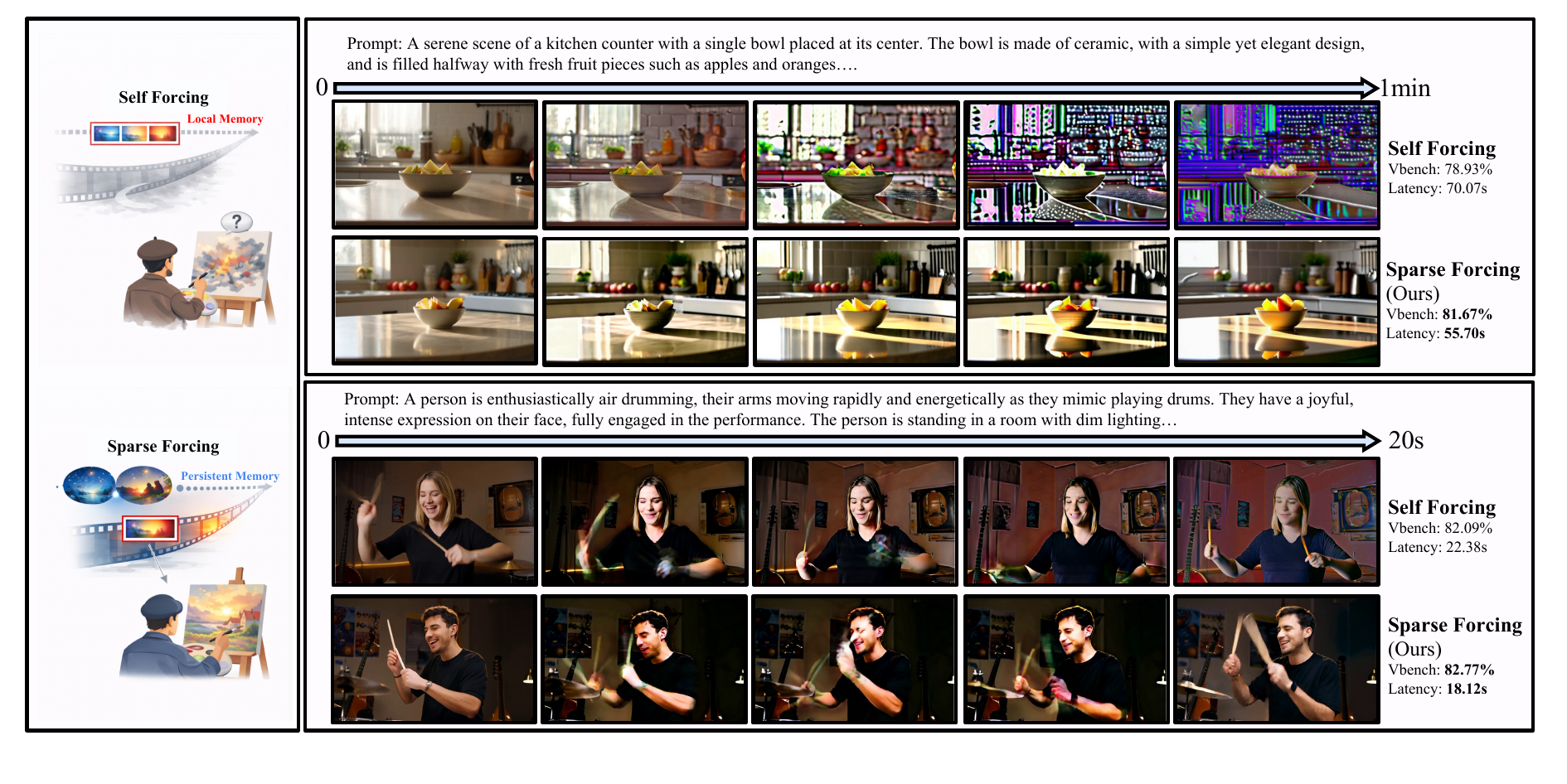}
    \caption{(Left) Illustration: Sparse Forcing keeps local context with persistent moments, preserving long-term generation stability with lower latency and alleviating drift and instability over time. (Right) Sparse Forcing leverages persistent spatiotemporal implicit memory and trainable native sparsity to improve long-horizon generation quality while reducing decoding latency.}
    \label{fig:Example}
\end{figure*}

The pursuit of high-fidelity multi-modal content generation has become a cornerstone of spatial intelligence and general-purpose AI, pushing the boundaries of how models perceive, predict, and simulate temporal dynamics. Among these modalities, video stands out as particularly demanding: it requires coherent long-range dynamics, fine-grained spatial fidelity, and efficient inference under rapidly growing context lengths.

Diffusion models have recently revolutionized text-to-video generation~\citep{blattmann2023stable, ho2022video, jinpyramidal, polyak2024movie, yangcogvideox, zheng2024open}. Many state-of-the-art systems adopt the Diffusion Transformer (DiT) architecture~\citep{bao2022all, peebles2023scalable}, which typically applies bidirectional attention across the full spatiotemporal token sequence. While effective, full-sequence attention incurs quadratic complexity in context length, making long-form video generation increasingly expensive in both latency and memory footprint.

To scale video diffusion to longer durations, autoregressive diffusion has emerged as a compelling alternative. Cauvid~\citep{yin2025slow} and self-forcing\citep{huang2025self} introduce a causal diffusion transformer with frame-wise dependencies, enabling sample-efficient training by leveraging supervision from all input frames at each iteration and accelerating inference via key-value (KV) caching, analogous to decoder-only large language models~\citep{brown2020language, radford2019language}. Autoregressive formulations are attractive for long-horizon synthesis and interactive settings, yet they introduce a fundamental challenge: during rollouts the model must condition on its own imperfect predictions, leading to compounding errors over time. Meanwhile, as the context grows, naively attending to all historical tokens remains computationally prohibitive.

A natural question is whether we can address quality and efficiency simultaneously. Existing efficiency techniques for video diffusion, such as quantization~\citep{zhaovidit} and few-step sampling or distillation~\citep{yin2024one, yin2024improved, kim2025autoregressive}, primarily aim to reduce computation across multiple diffusion steps. 
Another promising direction is to exploit sparsity. Recent works explore sparse video generation and sparse attention mechanisms~\citep{xisparse, zhang2025faster}, as well as structured designs such as sliding or tiled attention. In language modeling, native sparse attention has been studied as a principled approach that can be trained end-to-end and yields test-time acceleration~\citep{yuan2025native}. However, for autoregressive video diffusion, it remains unclear how to design sparsity that is both trainable and rollout-compatible, and whether such sparsity can improve long-horizon quality beyond merely saving memory and compute.

Importantly, sparsity is not only a computational optimization. Under autoregressive rollouts, the attention pattern reshapes the dependency graph through which prediction errors propagate over time. Dense attention provides abundant pathways for early mistakes to influence future frames, amplifying compounding errors. In contrast, a structured sparse conditioning mechanism can control the \emph{topology} and \emph{gain} of error propagation by limiting how far and how strongly uncertain tokens can affect subsequent generations. This perspective argues that well-designed sparsity can improve generation quality and inference efficiency in a coupled manner.

Motivated by this insight, we begin with an empirical observation: autoregressive diffusion rollouts exhibit a strong \emph{persistent clustering} effect, where a compact subset of visual tokens persistently captures salient blocks across time, forming an \emph{implicit spatiotemporal memory}. Building on this structure, we propose \textbf{Sparse Forcing}, a novel trainable sparse attention for autoregressive video diffusion models. Sparse Forcing learns to \emph{compress, preserve, and update} persistent clustered blocks while restricting local computation to a compact dynamically-selected neighborhood. 
To make the approach practical at scale for both training and inference, we further develop \textbf{Persistent Block-Sparse Attention (PBSA)}, an efficient GPU kernel that accelerates sparse attention and memory updates for low-latency and memory-efficient decoding.
\begin{quote}\small
\emph{``We do not remember days, we remember moments.''}\footnote{Cesare Pavese, \emph{This Business of Living: Diaries 1935--1950}.} \emph{Time cannot be carried in full; we move through it with only a few sparsely luminous points in spacetime.}
\end{quote}
Our main contributions are threefold.
(1) We identify an empirical phenomenon in long-horizon autoregressive video diffusion models: blockified tokens exhibit strong spatiotemporal persistence in KV cache, yet are discarded by naive recency-based conditioning, leading to compounding errors during rollouts.
(2) We introduce Sparse Forcing, a trainable native sparsity paradigm that leverages persistent spatiotemporal implicit memory and block-structured sparse attention to simultaneously improve long-horizon generation quality and reduce decoding latency. Sparse Forcing consistently improves short-video generation quality, and its gains persist when scaling to long, minute-level videos.
(3) To make the approach practical and efficient, we develop Persistent Block-Sparse Attention (PBSA) kernels that accelerate sparse attention and memory updates, delivering end-to-end speedups for both training and inference with reduced memory footprint.

\section{Related Work}
\textbf{Bidirectional Video Generation Models.}
Video generation has advanced rapidly in recent years, with modern approaches mostly adopting the paradigms of denoising diffusion. Video diffusion has been explored in both pixel space\citep{ho2022imagen,singer2022make} and latent space\citep{blattmann2023stable}, with architectures evolving from U-Nets\citep{rombach2022high, hongcogvideo} to DiTs\citep{peebles2023scalable, gupta2024photorealistic}. 
Significant multi-billion parameter industrial investment has driven the development, including open-sourced models\citep{wan2025wan,kong2024hunyuanvideo} and closed-source models\citep{polyak2024movie, brooks2024video}. These models operate bidirectionally: each frame can attend to both past and future frames during denoising. While this bidirectional context enables high-quality synthesis for offline generation, it is incompatible with the causality required for real-time video generation.

\textbf{Autoregressive Diffusion Models and Long Video Generation.} 
Diffusion has become the driving force behind video synthesis, where a central challenge is length scaling. Training-free length-extension methods [43, 44, 48, 49, 84] reschedule noise or re-balance temporal frequency to stretch pretrained models beyond their training horizon. 
A complementary thread blends diffusion with causal prediction: Diffusion Forcing\citep{chen2024diffusion} and HistoryGuidance\citep{songhistory} enable variable horizon conditioning and stable long rollouts by noise injection. These approaches are adapted in industrial systems such as SkyReels-V2 \citep{chen2025skyreels} and MAGI-1\citep{teng2025magi}.
StreamDiT\citep{kodaira2025streamdit} combines multi-step distillation with a moving frame buffer and mixed partition training to generate results in real-time. To mitigate error accumulation during AR generation, Self-Forcing\citep{huang2025self} simulates AR rollout during training, while its extensions\citep{cui2025self,liu2025rolling,yang2025longlive} further improve length generalization.

\textbf{Efficient Video Generation.}
As we scale video generation to long horizons, large context windows become a bottleneck, driving a wave of efficient computational designs. Kernel advances such as FlashAttention\citep{dao2022flashattention, daoflashattention} improve throughput.

Another line of work leverages compressing the latent space or token sequence: token merging\citep{wu2025importance} and patch scaling \citep{lee2024video}, compact/variablerate tokenizers \citep{bachmann2025flextok}, highly compressed latent space\citep{hacohen2024ltx}, or multiscale pyramids with re-noising\citep{jinpyramidal}. 
Meanwhile, linear attention\citep{katharopoulos2020transformers} is also widely used in video generation, SANA\citep{xie2024sana, xiesana} introduced linear attention diffusion transformers, while SANA-Video \citep{chen2025sana} further extends this with block-linear attention and constant-size KV cache on video generation.

\textbf{Sparse Attention and its Nativity.}
There has been a lot of research and discussion about the useful of sparse attention for Large Language Models (LLMs). Sparse attention for LLMs falls into two categories: memory-efficient and compute-efficient. Memory-efficient methods\citep{xiaoefficient,xiaoduoattention, zhang2023h2o, tangquest, liu2023scissorhands} reduce memory load to accelerate decoding. Compute-efficient methods \citep{jiang2024minference,xiao2024infllm,
han2024lm, li2025mminference} focus on processing only critical tokens. 

As the long context, the exploration of sparse attention mechanisms for accelerating DiTs fall into two categories: static and dynamic, depending on whether to select critical tokens dynamically during runtime or statically offline. Static methods\citep{xisparse, zhangfast} predefine sparse patterns offline, such as identifying recent tokens as critical. These methods lack adaptability to diverse sparsity patterns, leading to suboptimal performance. Dynamic methods\citep{zhang2025spargeattn, yang2025sparse,xuxattention,Xia_2025_ICCV,zhang2025training, zhang2025faster} determine sparse patterns at runtime, selecting critical tokens through an additional identification step. However, on one hand, the prior work still focuses on  fixed-length (e.g., 5-second) video generation and don't consider the unique memory tracing in visual autoregressive models for video generation and have no corresponding optimization in KV caching.
The native feature of large-scale models have been explored\citep{yuan2025native}, demonstrates that sparse attention not only show better hardware efficiency but also have better generation quality. However, the sparsity emergence for such a complex dynamic memory system under long-context video generation scenario is unexplored but very important.

\section{Methodology}

\begin{figure*}
    \centering
    \includegraphics[width=\linewidth, trim=5mm 80mm 60mm 55mm, clip]{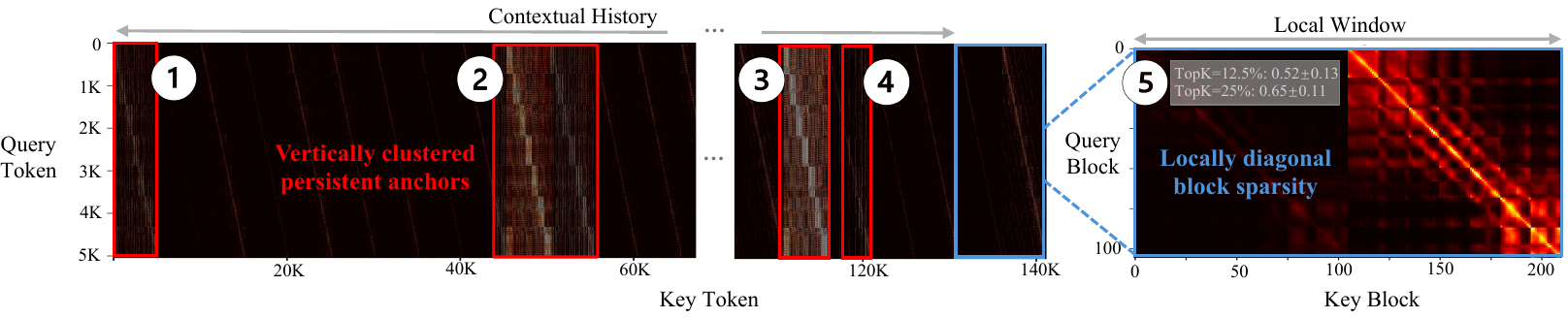}
    \caption{\textbf{Vertically clustered persistent anchors and local diagonal block sparsity.} Across the contextual history in self-forcing\citep{huang2025self}, attention concentrates on a few vertically clustered persistent anchors (\circnum{1}--\circnum{4}), whereas the local window exhibits a locally diagonal block-sparse pattern (\circnum{5}). Attention recall at $\text{Top-}K$=25\% is $0.65\pm0.11$ (mean$\pm$std over heads and layers).}
    \label{fig:persistency_and_block_attention}
\end{figure*}

\subsection{Autoregressive Video Diffusion Models}
Autoregressive video diffusion models combine sequential factorization with diffusion-based conditional generation. Instead of modeling a full video jointly, they produce it step by step, where each prediction depends on the previously generated context. Formally, for a video sequence $x_{1:N} = (x_1, x_2, \ldots, x_N)$, the distribution can be written as
\begin{equation}
p(x_{1:N}) \;=\; \prod_{i=1}^{N} p(x_i \mid x_{<i}) .
\end{equation}
Each conditional term $p(x_i \mid x_{<i})$ is implemented with a diffusion generator: the next frame is sampled by progressively denoising Gaussian noise while conditioning on preceding frames. In practice, one autoregressive step may generate a short chunk of consecutive frames rather than a single frame; for simplicity, we refer to that prediction unit as a ``frame'' throughout the paper.

There are two common ways to train such models: learning the autoregressive diffusion model directly from data, or distilling it from a pretrained bidirectional video diffusion model. In the former case, training usually follows either Teacher Forcing (TF)\citep{gaoca2, jinpyramidal, zhang2025test} or Diffusion Forcing (DF)\citep{chen2024diffusion, gu2025long}. With TF, the model conditions on clean ground-truth history frames. With DF, it still conditions on ground-truth history, but each historical frame is perturbed with an independently sampled noise level. While these strategies make optimization easier, they also create a mismatch between training and inference: during training the history is oracle-provided, whereas at test time the model must condition on its own past predictions. This mismatch is commonly referred to as exposure bias \citep{schmidt2019generalization}.

Reducing exposure bias in autoregressive diffusion is challenging because the denoising objective would ideally require supervision under the model's own sampled rollout states, and such paired targets are generally unavailable. Recent approaches therefore try to narrow the train--test gap by explicitly modeling rollout conditions during training or otherwise improving robustness to self-generated history\citep{yin2025slow, huang2025self}.

\subsection{Observation: Emergent Clustered Persistency and Local Block Sparse Attention}\label{sec:observation}

\begin{figure*}
    \centering
    \includegraphics[width=0.97\linewidth, trim=120mm 245mm 85mm 50mm, clip]{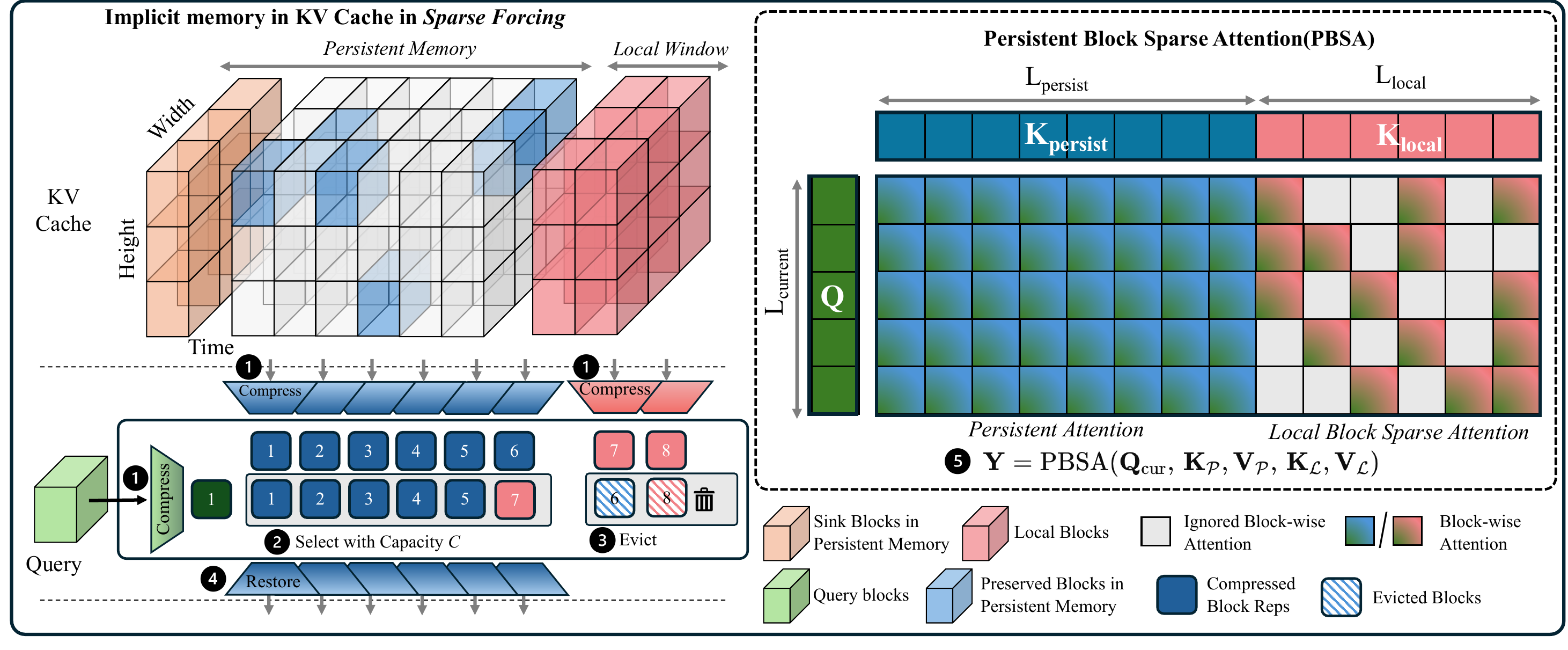}
    \caption{Overview of Sparse Forcing.}
    \label{fig:overview}
\end{figure*}

We observe two consistent attention patterns in autoregressive video diffusion rollouts that motivate Sparse Forcing.
First, \emph{emergent persistency}: over long horizons, attention concentrates on a small subset of historical blocks, forming vertically clustered persistent anchors that carry global context such as subject identity and scene layout.
Second, \emph{locally diverse block sparsity}: even within the recent window, attention allocation is highly structured and content-dependent, exhibiting a locally diagonal block-sparse pattern, as shown in \Cref{fig:persistency_and_block_attention}.
These patterns suggest that long-horizon generation can benefit from retaining a compact set of persistent anchors while sparsifying attention within the local window.

\textbf{Why Selective Preserving is Necessary.}
Full KV caching over the entire past context does not scale to the current long-horizon autoregressive video generation.
For a 1.3B model generating a 1-minute video, the FP16 KV cache reaches 44.9\,GB, i.e., $17.26\times$ the parameter memory even at batch size 1. This necessitates a memory-bounded yet effective KV selective preserving mechanism for long-horizon rollouts.

\textbf{Vertically Clustered Persistent Anchors.}
\Cref{fig:persistency_and_block_attention} shows \emph{vertically clustered} anchor regions in the contextual history (\circnum{1}--\circnum{4}): a small number of historical columns consistently receive substantial attention mass, while most past blocks contribute marginally.
This observation indicates that preserving a limited anchor block set may be sufficient for stability under long-horizon rollouts.

\textbf{Diverse Local Block Sparsity.}
Zooming into the local window in \Cref{fig:persistency_and_block_attention}, attention exhibits \emph{locally diagonal} block sparsity (\circnum{5}), reflecting structured short-term dependencies and leaving room to remove redundant dense computation.
Using block-level scoring to select $\text{Top-}K$ blocks, we achieve an attention token recall of $0.65\pm0.11$ at $\text{Top-}K$=25\% over heads and layers, suggesting that a moderate budget already recovers most important blocks.

\subsection{Autoregressive Video Generation in Sparse Forcing}

\textbf{A Structured Memory Decomposition.}
At autoregressive step $t$ and diffusion timestep $k$, Sparse Forcing maintains an implicit memory in KV caches:
\begin{equation}
\mathcal{M}_t^k = \mathcal{P}_t \cup \mathcal{L}_t^k,
\end{equation}
where $\mathcal{P}_t$ is a discrete set of persistent and fully denoised spatiotemporal blocks extracted at $k\text{=0}$ that carry long-range semantic anchors at $t$, and $\mathcal{L}_t^k$ is a local window containing the most recent spatiotemporally-contiguous blocks. 
$\mathcal{P}_t$ is shared across diffusion steps $k$ to provide stable long-range anchors, and is updated by coarse-grained scoring over compressed block representations, whereas $\mathcal{L}_t^k$ is updated by a sliding window as $t$ advances, and partially refreshed at each $(t,k)$ by only updating current denoising blocks.
Based on the emergence of persistence and block clustering discussed in \Cref{sec:observation}, for $\mathcal{P}_t$, we treat evicted blocks from $\mathcal{L}_t^k$ as candidates and maintain $\mathcal{P}_t$ via $\text{Top-}C$ retention.

\textbf{Persistent Block Sparse Attention (PBSA) in Sparse Forcing.}
Sparse Forcing maintains a bounded KV memory consisting of a persistent set of spatiotemporal blocks and a streaming local window, as shown in \Cref{fig:overview}.

Concretely, at each autoregressive step we keep a persistent memory $\mathcal{P}_t$ with capacity $|\mathcal{P}_t|\le C$,
which includes (i) sink blocks $\mathcal{S}_t$ as global anchors and (ii) a dynamic subset $\mathcal{D}_t$ selected from generation history, together with a local window $\mathcal{L}_t$ that stores the most recent blocks.

Given current queries $\mathbf{Q}_{\mathrm{cur}}\in\mathbb{R}^{N_q\times d}$,
persistent keys/values $(\mathbf{K}_{\mathcal{P}},\mathbf{V}_{\mathcal{P}})\in\mathbb{R}^{N_p\times d}$,
and local keys/values $(\mathbf{K}_{\mathcal{L}},\mathbf{V}_{\mathcal{L}})\in\mathbb{R}^{N_\ell\times d}$,
PBSA computes a \emph{single} masked attention over concatenated keys/values:
\begin{equation}
\mathbf{K}=[\mathbf{K}_{\mathcal{P}};\mathbf{K}_{\mathcal{L}}], 
\mathbf{V}=[\mathbf{V}_{\mathcal{P}};\mathbf{V}_{\mathcal{L}}]
\end{equation}
\begin{equation}
\mathbf{Y}_{\mathrm{cur}}
=
\mathrm{Softmax}\!\left(\frac{\mathbf{Q}_{\mathrm{cur}}\mathbf{K}^{\top}}{\sqrt d}+\mathbf{M}\right)\mathbf{V}
\label{eq:pbsa}
\end{equation}
The mask $\mathbf{M}\in\mathbb{R}^{N_q\times (N_p+N_\ell)}$ enforces \emph{dense} access to persistent anchors and
\emph{block-sparse} access within the local window:
\begin{equation}
\mathbf{M}=\big[\mathbf{0}_{N_q\times N_p}\; \mathbf{M}_{\mathcal{L}}\big],
\mathbf{M}_{\mathcal{L}}[q,\ell] = \log \mathbb{I}\!\left[\ell\in\Omega(q)\right]
\end{equation}
where $\log \mathbb{I}[\cdot]$ equals $0$ for visible blocks and $-\infty$ otherwise, and $\Omega(q)$ specifies visible local blocks for query block $q$.

\textbf{Blockified Compression.}
To enable efficient block-level scoring for maintaining the persistent set $\mathcal{P}_t$ while preserving spatiotemporal locality, we first \emph{blockify} the latent into contiguous blocks and then compute compact block representatives.

\emph{Blockify and locality-preserving layout.}
Given a latent tensor $\mathbf{X}\in\mathbb{R}^{T\times H\times W\times d}$,
we partition it into spatiotemporal blocks of size $(B_t,B_h,B_w)$, where each block contains $B=B_tB_hB_w$ tokens.
This induces a two-level indexing: \emph{block indices} $(t_b,h_b,w_b)$ and \emph{in-block indices} $(\Delta t,\Delta h,\Delta w)$.
We then reshape and permute $\mathbf{X}$ into a block-contiguous layout $\mathbf{X}^{\mathrm{blk}}\in\mathbb{R}^{N_b\times B\times d}$, as elaborated in \Cref{app:locality_preserving},
so that tokens within each block are stored contiguously in memory, enabling coalesced access and efficient block-level routing.

\emph{Block representatives.}
We use the superscript $(\cdot)^{c}$ to denote \emph{compressed} block representatives. On top of $\mathbf{X}^{\mathrm{blk}}$, we compute compact block representatives for both queries and keys:
\begin{equation}
\mathbf{Q}^{c}_{t} = \phi_{\mathbf{Q}}(\mathbf{Q}^{\mathrm{blk}}_{t}),
\mathbf{K}^{c}_{:t} = \phi_{\mathbf{K}}(\mathbf{K}^{\mathrm{blk}}_{:t})
\end{equation}
Here $\phi_{\mathbf{Q}}(\cdot)$ and $\phi_{\mathbf{K}}(\cdot)$ operators compress all $B$ tokens in a block into a single representative, such as pooling,
reducing the sequence length by a factor of $B$. As a result, $\mathbf{Q}^{c}_{t} \in \mathbb{R}^{N_q^{\mathrm{blk}}\times d}$ and $\mathbf{K}^{c}_{:t} \in \mathbb{R}^{N_k^{\mathrm{blk}}\times d}$, where $N_q^{\mathrm{blk}}=N_q/B$ and $N_k^{\mathrm{blk}}=N_k/B$. These block representatives are used only for coarse scoring and masking, while fine-grained attention operates on unmasked tokens.

\textbf{Coarse Scoring and Top-$C$ Persistent Update.}
Using the compressed block representatives, we perform coarse routing to estimate the long-range relevance of historical blocks to the current generation step.
Specifically, we compute a block-level attention matrix
\begin{equation}
\mathbf{A}_t
=
\mathrm{Softmax}\!\left(
\frac{\mathbf{Q}^{c}_{t}\left(\mathbf{K}^{c}_{:t}\right)^\top}{\sqrt d}
\right)
\in \mathbb{R}^{N_q^{c}\times N_k^{c}}
\label{eq:coarse_attn}
\end{equation}
where $\mathbf{A}_t[i,j]$ measures how much the $i$-th \emph{query block} attends to the $j$-th \emph{key block} at the coarse level.
To obtain a single importance score per key block, we aggregate attention weights across all query blocks:
\begin{equation}
\mathbf{s}_t=\frac{1}{N_q^{c}}\sum_{i=1}^{N_q^{c}}\mathbf{A}_t[i,:]
\label{eq:block_score}
\end{equation}
yielding $\mathbf{s}_t\in\mathbb{R}^{N_k^{c}}$ that ranks historical blocks by their relevance to the current step.

We maintain a bounded persistent memory by applying Top-$C$ retention over candidate blocks.
Let $\mathcal{E}_t$ denote the set of candidate blocks recently evicted from the local window $\mathcal{L}_t$.
We update the persistent set by
\begin{equation}
\mathcal{P}_t
=
\mathrm{Top}\text{-}C\big(\mathcal{P}_{t-1}\cup\mathcal{E}_t;\; \mathbf{s}_t\big),
\quad
|\mathcal{P}_t|\le C
\label{eq:topc_update}
\end{equation}
where $\mathrm{Top}\text{-}C(\cdot)$ retains the $C$ blocks with the highest aggregated scores.\footnote{Sink blocks are always retained as global anchors and are excluded from eviction.}
This demand-driven update promotes blocks that consistently receive high coarse relevance to become long-range anchors, while evicting less useful history to enforce a fixed memory budget.

\begin{algorithm}[t]
\caption{Autoregressive Diffusion Inference in Sparse Forcing}
\small
\label{alg:sparse_forcing_infer}
\begin{algorithmic}[1]
\REQUIRE Local window size $L_{\text{local}}$, $\text{Top-}K$  
\REQUIRE Persistent Memory Capacity $C$
\REQUIRE Denoise timesteps $\{t_1, \ldots, t_T\}$
\REQUIRE Number of generated frames $M$
\REQUIRE AR diffusion model $G_{\theta}$ (updates KV via $G^{KV}_{\theta}$)

\STATE Initialize model output $X_{\theta} \leftarrow [\,]$
\STATE Initialize persistent memory $\mathcal{P} \leftarrow [\,]$ 
\COMMENT Capacity $C$
\STATE Initialize local window $\mathcal{L} \leftarrow [\,]$
\COMMENT Capacity $L_{\text{local}}$
\FOR{$i = 1, \ldots, M$}
  \STATE Initialize $x^{i}_{t_T} \sim \mathcal{N}(0, I)$
  \FOR{$j = T, \ldots, 1$}
    \STATE Set $\hat{x}^{i}_{0} \leftarrow G_{\theta}(x^{i}_{t_j}; t_j, \mathcal{P}, \mathcal{L})$
    \COMMENT Apply PBSA with $\text{Top-}K$
    \IF{$j = 1$}
        \STATE $X_{\theta}.\mathrm{append}(\hat{x}^{i}_{0})$
        \STATE $(\mathcal{P},\mathcal{L}) \leftarrow G^{KV}_{\theta}(\hat{x}^{i}_{0}; 0, \mathcal{P}, \mathcal{L})$
        \COMMENT{updates $\mathcal{P},\mathcal{L}$}
    \ELSE
      \STATE Sample $\epsilon \sim \mathcal{N}(0, I)$
      \STATE Set $x^{i}_{t_{j-1}} \leftarrow \Psi(\hat{x}^{i}_{0}, \epsilon, t_{j-1})$
    \ENDIF
  \ENDFOR
\ENDFOR
\STATE \textbf{return} $X_{\theta}$
\end{algorithmic}
\end{algorithm}


\newcommand{\SR}{\cellcolor{SparseRow}}

\begin{table*}[t]
  \centering
  \caption{Comparison with relevant baselines. We compare Sparse Forcing with representative open-source video generation models of comparable scale and resolution. Best results are in \textbf{bold} and second-best results are \underline{underlined}.
  $^{\blacklozenge}$: with pretraining; $^{\diamond}$: without pretraining;
  $^{\spadesuit}$: [3,4,4] block size; $^{\clubsuit}$: [1,8,8] block size.}
  \label{tab:short_video_comparison_baselines}
  \small
  \footnotesize
  \setlength{\tabcolsep}{4.2pt}
  \renewcommand{\arraystretch}{1.08}

  \begin{tabularx}{\textwidth}{>{\raggedright\arraybackslash}X c c c c c c c}
    \toprule
    \multirow{2}{*}{Model} &
    \multirow{2}{*}{\#Params} &
    \multirow{2}{*}{Resolution} &
    \multirow{2}{*}{\makecell[c]{Throughput\\(FPS) $\uparrow$}} &
    \multirow{2}{*}{\makecell[c]{Latency\\(s) $\downarrow$}} &
    \multicolumn{3}{c}{Evaluation scores $\uparrow$} \\
    \cmidrule(lr){6-8}
    & & & & & Total & Quality & Semantic \\
    \midrule

    \multicolumn{8}{l}{\textbf{Diffusion models}} \\
    LTX-Video\cite{hacohen2024ltx} & 1.9B & $768{\times}512$ & 8.98 & 13.5 & 80.00 & 82.30 & 70.79 \\
    Wan2.1\cite{wan2025wan}        & 1.3B & $832{\times}480$ & 0.78 & 103  & 84.26 & 85.30 & 80.09 \\
    \midrule

    \multicolumn{8}{l}{\textbf{Chunk-wise autoregressive models}} \\
    SkyReels-V2\cite{chen2025skyreels} & 1.3B & $960{\times}540$ & 0.49 & 112 & 82.67 & 84.70 & 74.53 \\
    MAGI-1\cite{teng2025magi}          & 4.5B & $832{\times}480$ & 0.19 & 282 & 79.18 & 82.04 & 67.74 \\

    CausVid\cite{yin2025slow}          & 1.3B & $896{\times}512$ & 17.0   & 0.69  & 82.69 & 83.73 & 78.49 \\

    Self Forcing\cite{huang2025self} &
    1.3B & $896{\times}512$ & 17.0 & 0.69 &
    83.88 & 84.60 & 81.01 \\

    \SR Sparse Forcing$^{\diamond\spadesuit}$ &
    \SR 1.3B & \SR $896{\times}512$ & \SR 19.9 & \SR 0.59 &
    \SR 83.99 & \SR 84.65 & \SR 81.36 \\

    \SR Sparse Forcing$^{\diamond\clubsuit}$ &
    \SR 1.3B & \SR $896{\times}512$ & \SR 18.8 & \SR 0.63 &
    \SR 83.91 & \SR 84.58 & \SR 81.24 \\
    
    \SR Sparse Forcing$^{\blacklozenge\spadesuit}$ &
    \SR 1.3B & \SR $896{\times}512$ & \SR 19.9 & \SR 0.59 &
    \SR \textbf{84.14} &
    \SR \textbf{84.84} &
    \SR \textbf{81.39} \\


    \bottomrule
  \end{tabularx}
\end{table*}

\textbf{Block-Sparse Attention in Local Window.}
We maintain a sliding local window $\mathcal{L}_t$ and apply block sparsity only within $\mathcal{L}_t$ to reduce computation while preserving recent details.

\emph{Row-wise Top-$K$ Block Selection.}
To instantiate the local sparsity pattern in \Cref{eq:pbsa}, we derive $\Omega(q)$ via lightweight coarse routing within $\mathcal{L}_t$.
Using compressed block representatives $(\mathbf{Q}^{c}_{\mathrm{cur}},\mathbf{K}^{c}_{\mathcal{L}})$, we compute
\begin{equation}
\mathbf{A}_{\mathcal{L}}
=
\mathrm{Softmax}\!\left(\frac{\mathbf{Q}^{c}_{\mathrm{cur}}(\mathbf{K}^{c}_{\mathcal{L}})^\top}{\sqrt{d}}\right)
\end{equation}

Then, we define the routed key-block set for each query block $q$ by row-wise $\text{Top-}K$ selection:
\begin{equation}
\Omega(q)=\operatorname{arg\,TopK}_{j}\ \mathbf{A}_{\mathcal{L}}[q,j],
\quad
|\Omega(q)|=N_\ell^{\text{blk}}\times K
\label{eq:rowwise_topk}
\end{equation}
This induces a block-sparse local mask $\mathbf{M}_{\mathcal{L}}$ while keeping $\mathcal{P}_t$ fully visible and densely attended by all queries in $\mathbf{Q}_{\mathrm{cur}}$.

\subsection{The Customized Kernels for Sparse Forcing}\label{method:kernel}
Sparse Forcing requires an efficient block-sparse attention kernel with \emph{persistent} implicit KV memory, supporting both forward and backward execution.
Most existing customized sparse-attention kernels for video generation are primarily developed for diffusion models, where the sparsity pattern is largely \emph{static}~\citep{li2025radial,xisparse} and the attention geometry is typically \emph{regular} (e.g., fixed sequence length and a square $QK^\top$ geometry with $L_q = L_k$)~\citep{zhang2025faster,yang2025sparse}.
In contrast, Sparse Forcing maintains a \emph{persistent} implicit memory that is carried across autoregressive steps with bounded capacity, while attending to a dynamically selected set of blocks within local windows conditioned on the current context.
Such a globally persistent and locally dynamic block sparse attention structure is not directly supported by prior GPU kernels, which typically assume limited sparsity layouts and lack primitives for persistent KV carry-over and selective updates.

To bridge this gap, we implement \textbf{Persistent Block-Sparse Attention (PBSA)} kernel using ThunderKittens~\citep{spector2024thunderkittens}, tailored for the coarse and fine stage of Sparse Forcing.
PBSA supports persistent-block carry-over, dynamic block selection over the non-persistent region, enabling efficient end-to-end training and inference.
Consequently, PBSA substantially reduces the runtime overhead of Sparse Forcing, making its training and inference practical at scale.

\subsection{Training for Sparse Forcing}
Sparse Forcing distills a pretrained bidirectional video diffusion model into a few-step causal autoregressive generator using the distribution matching distillation (DMD) loss. 
The full training procedure is provided in \Cref{app:training_alg}.
A training-free application of Sparse Forcing at inference introduces a train--test mismatch: 
the base model is optimized under a non-rolling cache and dense local attention assumption, whereas decoding operates with a dynamic cache and adaptive local attention, which can amplify compounding errors and manifest as visual artifacts.
To close this gap, we enable dynamically updated cache and adaptive local attention during training, faithfully mitigating the mismatch and stabilizing rollouts beyond the training horizon.



\begin{table}[t]
  \centering
  \caption{Comparison with baselines on long-horizon generation.
  $^{\blacklozenge}$: with pretraining; 
  $^{\spadesuit}$: [3,4,4] block size; $^{\clubsuit}$: [1,8,8] block size.}
  \label{tab:long_video_comparison_baselines}
  \small
  \setlength{\tabcolsep}{3.2pt}
  \renewcommand{\arraystretch}{1.06}

  \begin{tabularx}{\columnwidth}{>{\raggedright\arraybackslash}X c c c}
    \toprule
    Model &
    \makecell[c]{FPS$\uparrow$} &
    \makecell[c]{Latency/s $\downarrow$} &
    \makecell[c]{VBench $\uparrow$(T/Q/S)} \\
    \midrule

    \multicolumn{4}{l}{\textit{20-second length video}} \\
    Self Forcing & 14.4 & 0.83 & 82.09 / 82.48 / 80.51 \\


    \SR Sparse Forcing$^{\blacklozenge\spadesuit}$ &
    \SR 18.3 & \SR 0.65 &
    \SR \textbf{82.68} / \textbf{83.13} / \textbf{80.87} \\


    \SR Sparse Forcing$^{\blacklozenge\clubsuit}$ &
    \SR 17.9 & \SR 0.67 &
    \SR \underline{82.31} / \underline{82.64} / \underline{81.01} \\

    \midrule
    \multicolumn{4}{l}{\textit{1-minute length video}} \\
    Self Forcing & 13.9 & 0.87 & 78.93 / 79.48 / 76.70 \\


    \SR Sparse Forcing$^{\blacklozenge \spadesuit}$ &
    \SR 18.0 & \SR 0.66 &
    \SR \textbf{81.96} / \textbf{82.25} / \textbf{80.82}  \\


    \SR Sparse Forcing$^{\diamond \clubsuit}$ &
    \SR 17.6 & \SR 0.66 &
    \SR \underline{81.67} / \underline{82.17} / \underline{79.67} \\


    \bottomrule
  \end{tabularx}
\end{table}

\section{Experiments}
\subsection{Training and Evaluation Settings} 
\textbf{Training.}
We train Sparse Forcing variants and baselines on 5-second video clips using 8$\times$ NVIDIA H100 GPUs.
We build Sparse Forcing on top of Wan2.1-T2V-1.3B~\citep{wan2025wan} as the base text-to-video diffusion model. Following CausVid~\citep{yin2025slow} and Self-Forcing~\citep{huang2025self}, we initialize the base model under a causal attention mask using 16K ODE solution pairs sampled from the base model. Then, we use 4-step diffusion sampling during training and perform chunk-wise denoising, where each chunk contains 3 temporal latent frames. 
We adopt distribution matching distillation (DMD) with text prompts drawn from a filtered and LLM-extended version of VidProM~\citep{wang2024vidprom}. We train Sparse Forcing for 1200 steps with a batch size of 64 using AdamW.
For memory compression, we use average pooling as the compression operator, and we set the persistent-memory capacity $C=6$ frames and the local-window length to $L_{\text{local}}=6$ frames as well. We set $\text{Top-}K=25\%$ for row-wise block sparse selection within local windows. 
The training dynamics and implementation details are in \Cref{app:training_dynamics,app:training_setting}.

\textbf{Evaluation.}
We evaluate 4,730 generated videos (946 prompts $\times$ 5 samples per prompt) using VBench~\citep{huang2024vbench} to assess both semantic alignment and perceptual quality. VBench reports 16 metrics including 9 semantic dimensions (e.g., spatial relationship and object class) and 7 perceptual-quality dimensions (e.g., aesthetic quality and imaging quality). Following~\citep{huang2025self}, we rewrite text prompts with Qwen2.5-7B-Instruct. We additionally observe that dynamic degree and color metrics are particularly prone to degradation in long-horizon generation.

\subsection{Quantitative Comparison}
We evaluate Sparse Forcing on both short- and long-horizon video generation.
Across all horizons, Sparse Forcing consistently improves generation quality while reducing peak memory and inference latency.
Notably, when scaling to 20-second and 1-minute videos, 4 to 12$\times$ longer rollouts than training, these gains hold without any extrapolation-specific optimization, yielding a dominant quality--efficiency profile over baselines.


\textbf{Evaluation on Short Video.}
On 5-second short clips, Self Forcing degenerates to full attention since the window covers the entire sequence.
\Cref{tab:short_video_comparison_baselines} shows that Sparse Forcing achieves higher quality across evaluation metrics with a faster decoding speed and $42\%$ less peak KV cache, compared with full attention. Notably, even when training-free, Sparse Forcing also improves generation quality over the baseline in a plug-and-play setting. A comprehensive evaluation of the VBench metrics is provided in \Cref{app:full_vbench_eval}.

\begin{figure*}
    \centering
    \includegraphics[width=\linewidth, trim=0mm 0mm 0mm 0mm, clip]{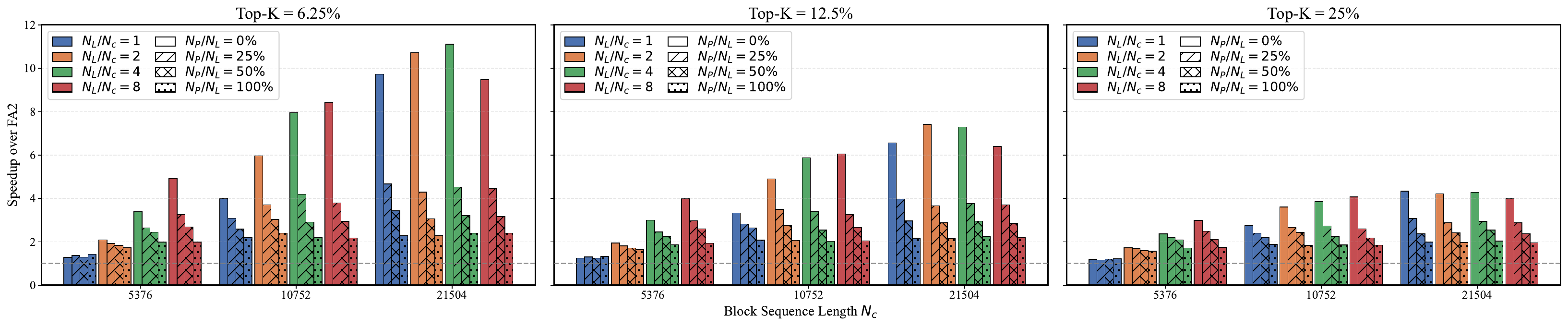}
    \caption{End-to-end speedup of PBSA over FA2.}
    \label{fig:kernel_speedup}
\end{figure*}

\textbf{When adapted onto Long videos.}
\Cref{tab:long_video_comparison_baselines} shows that, when extended to long-video generation, Sparse Forcing consistently outperforms the baseline in both semantic alignment and overall video quality,  while achieving higher throughput and lower peak memory. This indicates favorable test-time scaling of Sparse Forcing beyond the training horizon. Additional generation examples are provided in \Cref{app:more_generation_samples}.


\subsection{Qualitative Comparison}
\begin{figure*}
    \centering
    \includegraphics[width=\linewidth, trim=0mm 0mm 0mm 0mm, clip]{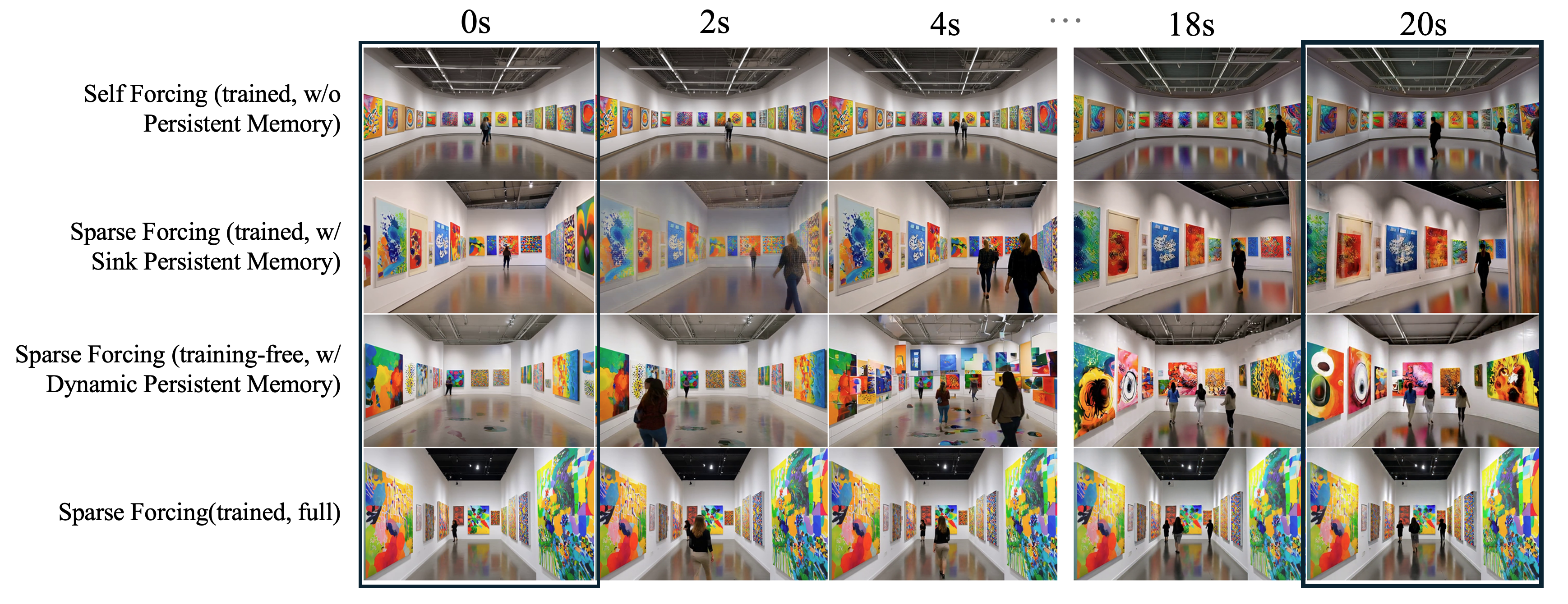}
    \caption{\textbf{Qualitative long-horizon rollouts in a gallery scene.} Sparse Forcing (trained, full) best preserves appearance consistency (color tone, stability, and identity consistency).}
    \label{fig:4_group_ablation}
\end{figure*}

\Cref{fig:Example} shows that Sparse Forcing sustains visually consistent long-horizon rollouts, largely preserving color tone and appearance coherence, whereas self forcing exhibits pronounced drift and accumulating artifacts. 
\Cref{fig:4_group_ablation} further diagnoses these failure modes and clarifies the roles of persistent memory and train--test alignment. Self forcing with a sliding window progressively compounds errors, leading to geometric distortion and color drift; 
Sparse Forcing using sink frames as persistent memory under the same KV budget, partially stabilizes the rollout, but flickering and ghosting remain.
Meanwhile, Sparse Forcing with training-free dynamic persistent memory can introduce semantic rewrites, revealing a mismatch between learned retrieval dynamics and the imposed memory update rule. 
In contrast, the full Sparse Forcing model learns to dynamically update memory during training, substantially improving long-horizon appearance consistency and temporal coherence. Additional examples are provided in \Cref{app:ablated_comparision}.

\subsection{Kernel Performance}
We benchmark an end-to-end execution latency of the PBSA kernel on NVIDIA H100 96GB GPUs, and report the speedup over FlashAttention-2 (FA2)~\citep{daoflashattention} in \Cref{fig:kernel_speedup}. 
Detailed evaluations of the forward and backward passes are provided in \Cref{app:PBSA_latency}.
\Cref{fig:kernel_speedup} brings a comprehensive understanding on the sparsity level (top-$k$ block ratio), the local-window expansion ratio ($N_L/N_c$), the persistent-to-local ratio ($N_P/N_L$), and the per-block sequence length $N_c$. Across all evaluated settings, PBSA achieves consistent speedups over FA2, ranging from $1.16\times$ to $11.11\times$.

\textbf{Impact of Sparsity, Local Window, and Persistence.}
PBSA yields larger gains under stronger local sparsity. As $\text{top-}K$ decreases from $25\%$ to $12.5\%$ and $6.25\%$, the peak speedup rises from $4.34\times$ to $7.29\times$ and $11.11\times$. Meanwhile, speedups further improve with longer sequences and larger local windows, where block-level sparsity better amortizes attention cost and memory movement.
Finally, PBSA is most effective when the persistent portion is compact, guiding effective hyperparameter settings in Sparse Forcing and further model design. A detailed latency breakdown is provided in \Cref{app:pbsa_latency_breakdown}.

\subsection{Ablation Study}
We disentangle the impact of three core components in Sparse Forcing: (i) maintaining a persistent memory $\mathcal{P}$, (ii) applying block-sparse attention within the local window $\mathcal{L}$, and (iii) with continuous pretraining. The results in \Cref{tab:ablation_dims_fps} show that each component contributes meaningfully to long-horizon quality and/or decoding throughput, while the full model achieves the best overall trade-off between VBench dimensions and throughput.

\begin{table}[t]
  \centering
  \small
  \caption{\textbf{Ablation Study on Sparse Forcing.} VBench dimension scores and throughput (FPS) on 20s generation. }
  \label{tab:ablation_dims_fps}
  \setlength{\tabcolsep}{3.6pt}
  \renewcommand{\arraystretch}{1.05}
  \begin{tabular}{@{}lccc@{}}
    \toprule
    \textbf{Method} &
    \makecell[c]{\textbf{Dynamic Degree}$\uparrow$} &
    \textbf{Color}$\uparrow$ &
    \textbf{FPS}$\uparrow$ \\
    \midrule
    Self Forcing
      & 56.67 & 82.07 & 14.4 \\
    \midrule
    \textbf{Sparse Forcing}
      & \textbf{66.39} & \textbf{89.47} & \underline{18.3} \\
    \quad w/o $\mathcal{P}$
      & 47.22 & 80.88 & \textbf{22.8} \\
    \quad w/o $\mathrm{BSA}$ in $\mathcal{L}$
      & 50.93 & 87.45 & 17.6 \\
    \quad w/o Cont.\ Pretrain
      & \underline{63.06} & \underline{87.68} & \underline{18.3} \\
    \bottomrule
  \end{tabular}
\end{table}







\section{Conclusion}
We provide an empirical characterization of long-horizon attention in autoregressive video diffusion rollouts, revealing emergent persistency and locally diagonal block sparsity. Guided by these findings, we propose Sparse Forcing, a trainable sparse-attention mechanism for autoregressive--diffusion hybrid video generation that improves long-range visual consistency while reducing decoding cost, together with an optimized kernel that supports irregular sparsity patterns for efficient training and deployment.
\newpage
\appendix
\onecolumn

\clearpage
\newpage
\bibliographystyle{assets/plainnat}
\bibliography{paper}

\clearpage
\newpage
\beginappendix

\section{Evaluation across training Steps}\label{app:training_dynamics}
\begin{figure*}[htbp!]
  \centering
  \includegraphics[
    width=\textwidth,
    height=0.2\textheight,
    keepaspectratio,
    trim=10mm 80mm 150mm 5mm, 
    clip
  ]{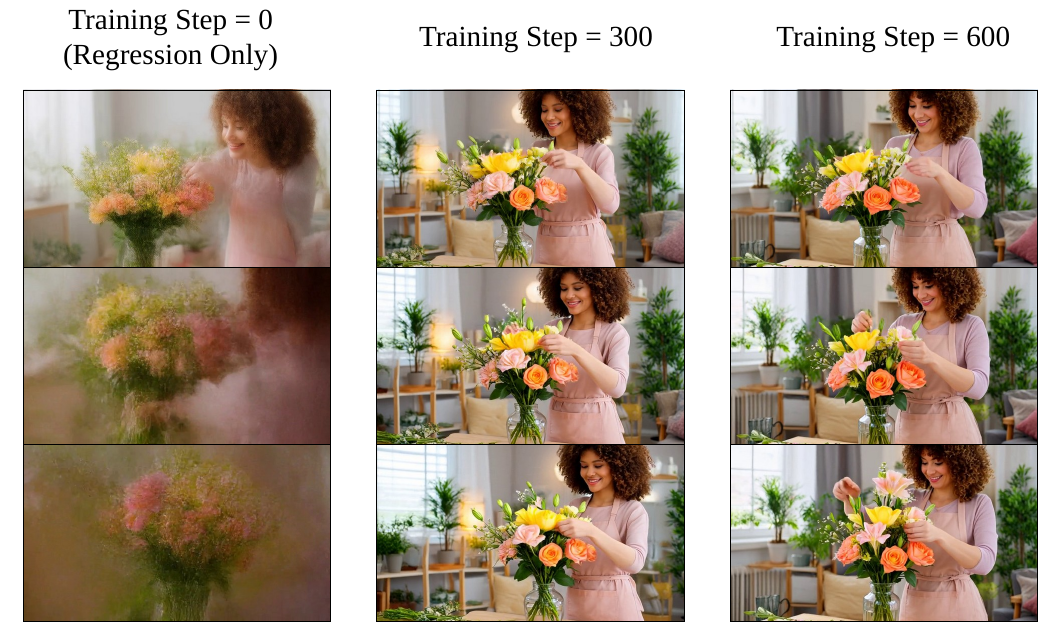}
  \includegraphics[
    width=\textwidth,
    height=0.2\textheight,
    keepaspectratio,
    trim=10mm 80mm 150mm 5mm,
    clip
  ]{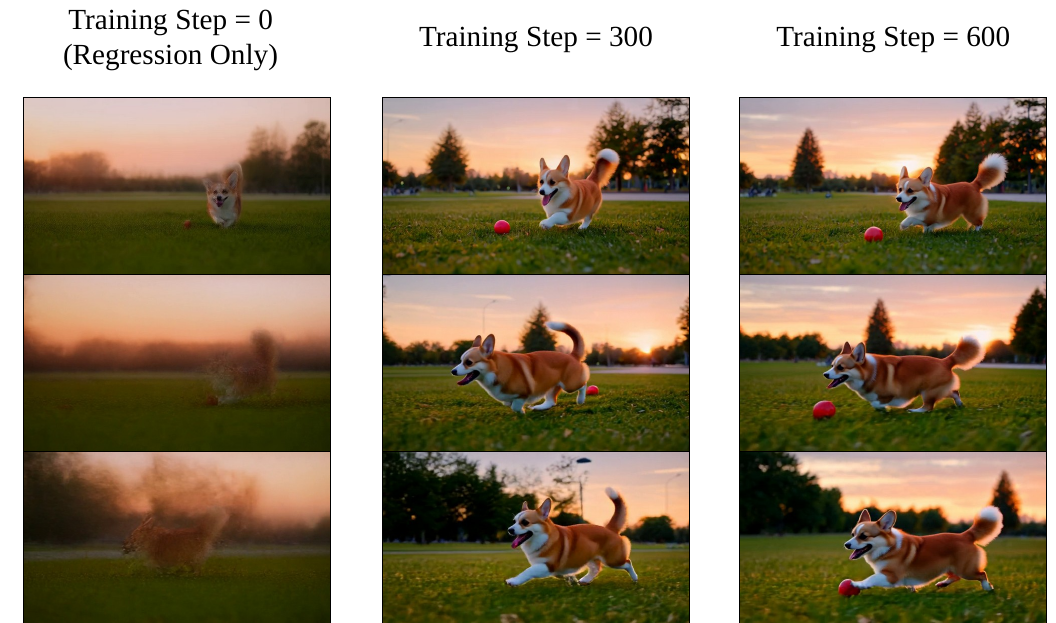}
  \includegraphics[
    width=\textwidth,
    height=0.2\textheight,
    keepaspectratio,
    trim=10mm 80mm 150mm 5mm,
    clip
  ]{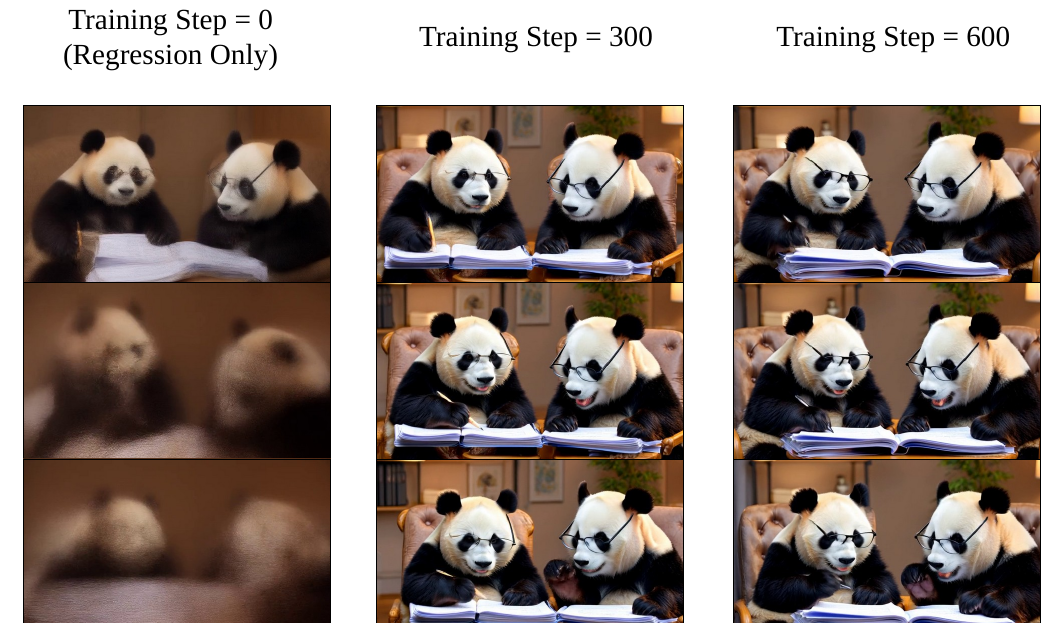}
  \caption{Qualitative results at different training steps on three video samples in Sparse Forcing.}
  \label{fig:train_dynamics}
\end{figure*}

\Cref{fig:train_dynamics} qualitatively visualizes how Sparse Forcing evolves over training, using three representative video prompts and three snapshots of the model state. 
At step~0 (regression-only), the model typically collapses to low-frequency, over-smoothed predictions, and temporal details are weakly grounded, resulting in blurred frames and unstable motion.
As causal distillation training proceeds, we observe a consistent coarse-to-fine refinement: by step~300, the model begins to recover object boundaries and salient appearance cues, while motion becomes more coherent across frames.
By step~600, Sparse Forcing produces sharp spatial details and temporally consistent dynamics across all three examples. 
This suggests that, even under the sparse-memory regime enforced during training, causal distillation still enables the model to gradually adapt and improve temporal modeling capability.

\section{Training Algorithm for Sparse Forcing}\label{app:training_alg}
The training algorithm is given in \Cref{alg:sparse_forcing_training}.
During training, we only enable gradient computation at a stochastic diffusion timestep to make training faster, following the training process in \citep{huang2025self}. 


\begin{algorithm}[t]
\caption{Sparse Forcing Training}
\begin{algorithmic}[1]
\REQUIRE Local window size $L_{\text{local}}$, $\text{Top-}K$  
\REQUIRE Persistent Memory Capacity $C$
\REQUIRE Denoise timesteps $\{t_1, \ldots, t_T\}$
\REQUIRE Number of video frames $N$
\REQUIRE AR diffusion model $G_{\theta}$ (Updates KV via $G^{KV}_{\theta}$)

\LOOP
  \STATE Initialize model output $X_{\theta} \leftarrow [\,]$
  \STATE Initialize persistent memory $\mathcal{P} \leftarrow [\,]$ \COMMENT{Capacity $C$}
  \STATE Initialize local window $\mathcal{L} \leftarrow [\,]$
  \COMMENT Capacity $L_{\text{local}}$
  \STATE Sample $s \sim \mathrm{Uniform}(1,2,\ldots,T)$
  \FOR{$i = 1, \ldots, N$}
    \STATE Initialize $x^{i}_{t_T} \sim \mathcal{N}(0, I)$
    \FOR{$j = T, \ldots, s$}
      \IF{$j = s$}
        \STATE Enable gradient comp.
        \STATE Set $\hat{x}^{i}_{0} \leftarrow G_{\theta}(x^{i}_{t_j}; t_j, \mathcal{P}, \mathcal{L})$
        \COMMENT Apply PBSA with $\text{Top-}K$
        \STATE $X_{\theta}.\mathrm{append}(\hat{x}^{i}_{0})$
        \STATE Disable gradient comp.
        \STATE Cache $\mathcal{P}, \mathcal{L} \leftarrow G^{KV}_{\theta}(\hat{x}^{i}_{0}; 0, \mathcal{P}, \mathcal{L})$
        \COMMENT Apply PBSA with $\text{Top-}K$ and Update $\mathcal{P}$ and $\mathcal{L}$
      \ELSE
        \STATE Disable gradient comp.
        \STATE Set $\hat{x}^{i}_{0} \leftarrow G_{\theta}(x^{i}_{t_j}; t_j, \mathcal{P}, \mathcal{L})$
        \COMMENT Apply PBSA with $\text{Top-}K$
        \STATE Sample $\epsilon \sim \mathcal{N}(0, I)$
        \STATE Set $x^{i}_{t_{j-1}} \leftarrow \Psi(\hat{x}^{i}_{0}, \epsilon, t_{j-1})$
      \ENDIF
    \ENDFOR
  \ENDFOR
  \STATE Update $\theta$ via Distribution matching distillation loss
\ENDLOOP
\end{algorithmic}
\label{alg:sparse_forcing_training}
\end{algorithm}

\section{Implementation Details}\label{app:training_setting}
\textbf{Training hyperparameters.}
\begin{table}[t]
\centering
\caption{Implementation details and training hyperparameters.}
\label{tab:training_hyperparameters}
\small
\setlength{\tabcolsep}{6pt}
\begin{tabular}{@{}l l@{}}
\toprule
\textbf{Item} & \textbf{Value} \\
\midrule
Real score network (DMD) & \texttt{Wan2.1-T2V-14B} \\
CFG weight & $3.0$ \\
Critic initialization & \texttt{Wan2.1-T2V-1.3B} \\
Batch size & $64$ \\
Optimizer ($G_{\theta}$) & AdamW \\
Optimizer ($f_{\psi}$) & AdamW \\
AdamW $\beta_1$ / $\beta_2$ & $0$ / $0.999$ \\
AdamW $\epsilon$ & $10^{-8}$ \\
Weight decay & $0.01$ \\
Learning rate ($G_{\theta}$) & $2\times 10^{-6}$ \\
Learning rate ($f_{\psi}$) & $4\times 10^{-7}$ \\
Generator/Critic Update ratio & $5{:}1$ \\
EMA decay & $0.99$ \\
\bottomrule
\end{tabular}
\vspace{-0.6em}
\end{table}
The training hyperparameters are listed in \Cref{tab:training_hyperparameters}.

\section{Full VBench Evaluations}\label{app:full_vbench_eval}
\Cref{tab:vbench_dim_breakdown} reports a full breakdown over all 16 VBench dimensions, comparing Sparse Forcing against the Self Forcing baseline across 5-second, 20-second, and 1-minute generation.
On 5-second short videos, Sparse Forcing generally improves semantic alignment and generation quality. It yields higher subject and background consistency, better overall consistency, and stronger temporal style.
Notably, the gains are most pronounced on semantics-heavy dimensions, including human action, object class, multiple objects, and scene, suggesting more faithful prompt-following and more stable object-level representations even at short horizons.

As the rollout length increases, the advantage of Sparse Forcing becomes more evident.
On 1-minute generation, Sparse Forcing substantially reduces long-horizon drift, improving subject/background consistency, and color fidelity, while also strengthening compositional metrics such as multiple objects and spatial relationship.
We also observe that motion-oriented scores such as temporal flickering and motion smoothness can be slightly lower for Sparse Forcing in some settings, which is consistent with a trade-off where maintaining richer dynamics may introduce mild temporal artifacts.
Overall, the full-metric evaluation supports that Sparse Forcing improves semantic correctness and long-horizon consistency, with the largest gains emerging as generation extends to the minute scale.

\begin{table*}[t]
  \centering
  \caption{\textbf{VBench Evaluation on different dimensions (\%) across generation lengths.} We compare \textsc{Self Forcing} and \textsc{Sparse Forcing} under 5-second, 20-second, and 1-minute generation. The better result within each pair is highlighted in bold.}
  \label{tab:vbench_dim_breakdown}
  \small
  \begin{tabular*}{\textwidth}{@{\extracolsep{\fill}} lcc cc cc}
    \toprule
    & \multicolumn{2}{c}{\textbf{5 seconds}} & \multicolumn{2}{c}{\textbf{20 seconds}} & \multicolumn{2}{c}{\textbf{1 minute}} \\
    \textbf{Dimension} &
    \shortstack{\textsc{Self}\\\textsc{Forcing}} &
    \shortstack{\textsc{Sparse}\\\textsc{Forcing}} &
    \shortstack{\textsc{Self}\\\textsc{Forcing}} &
    \shortstack{\textsc{Sparse}\\\textsc{Forcing}} &
    \shortstack{\textsc{Self}\\\textsc{Forcing}} &
    \shortstack{\textsc{Sparse}\\\textsc{Forcing}} \\
    \midrule
    subject consistency$\uparrow$        & 94.82 & \textbf{96.19} & 91.52 & \textbf{93.12} & 85.48 & \textbf{91.52} \\
    background consistency$\uparrow$     & 95.80 & \textbf{96.77} & 93.12 & \textbf{94.13} & 88.12 & \textbf{92.68} \\
    temporal flickering$\uparrow$        & 98.84 & \textbf{99.12} & \textbf{98.81} & 98.07 & \textbf{98.81} & 97.71 \\
    motion smoothness$\uparrow$          & \textbf{98.41} & 98.10 & \textbf{98.35} & 97.62 & \textbf{98.31} & 97.50 \\
    dynamic degree$\uparrow$             & \textbf{69.44} & 61.11 & 56.67 & \textbf{66.39} & 56.39 & \textbf{69.17} \\
    aesthetic quality$\uparrow$          & 67.15 & \textbf{67.90} & 65.72 & \textbf{66.22} & 61.86 & \textbf{64.71} \\
    imaging quality$\uparrow$            & \textbf{70.75} & 69.85 & 69.37 & \textbf{69.49} & 67.84 & \textbf{68.58} \\
    overall consistency$\uparrow$        & 25.42 & \textbf{26.74} & \textbf{27.14} & 26.89 & 26.63 & \textbf{26.90} \\
    temporal style$\uparrow$             & 22.73 & \textbf{24.15} & \textbf{24.56} & 24.43 & 24.32 & \textbf{24.52} \\
    human action$\uparrow$               & 80.80 & \textbf{97.00} & \textbf{96.20} & 95.60 & 95.40 & \textbf{95.60} \\
    object class$\uparrow$               & 88.49 & \textbf{95.65} & \textbf{94.32} & 93.59 & 89.08 & \textbf{93.07} \\
    multiple objects$\uparrow$           & 74.70 & \textbf{88.14} & 85.03 & \textbf{86.80} & 76.77 & \textbf{83.60} \\
    scene$\uparrow$                       & 44.97 & \textbf{56.72} & 55.81 & \textbf{57.53} & \textbf{54.99} & 54.29 \\
    appearance style$\uparrow$           & \textbf{20.62} & 20.54 & \textbf{20.96} & 20.66 & \textbf{21.20} & 20.89 \\
    color$\uparrow$                       & 89.81 & \textbf{89.87} & 82.07 & \textbf{89.47} & 71.70 & \textbf{86.54} \\
    spatial relationship$\uparrow$       & 77.10 & \textbf{81.27} & \textbf{83.74} & 79.23 & 76.38 & \textbf{77.91} \\
    \bottomrule
  \end{tabular*}
  \vspace{-3mm}
\end{table*}

\section{Ablated Comparison}\label{app:ablated_comparision}
\begin{figure*}
    \centering
    \includegraphics[width=\linewidth, trim=10mm 0mm 10mm 5.2mm, clip]{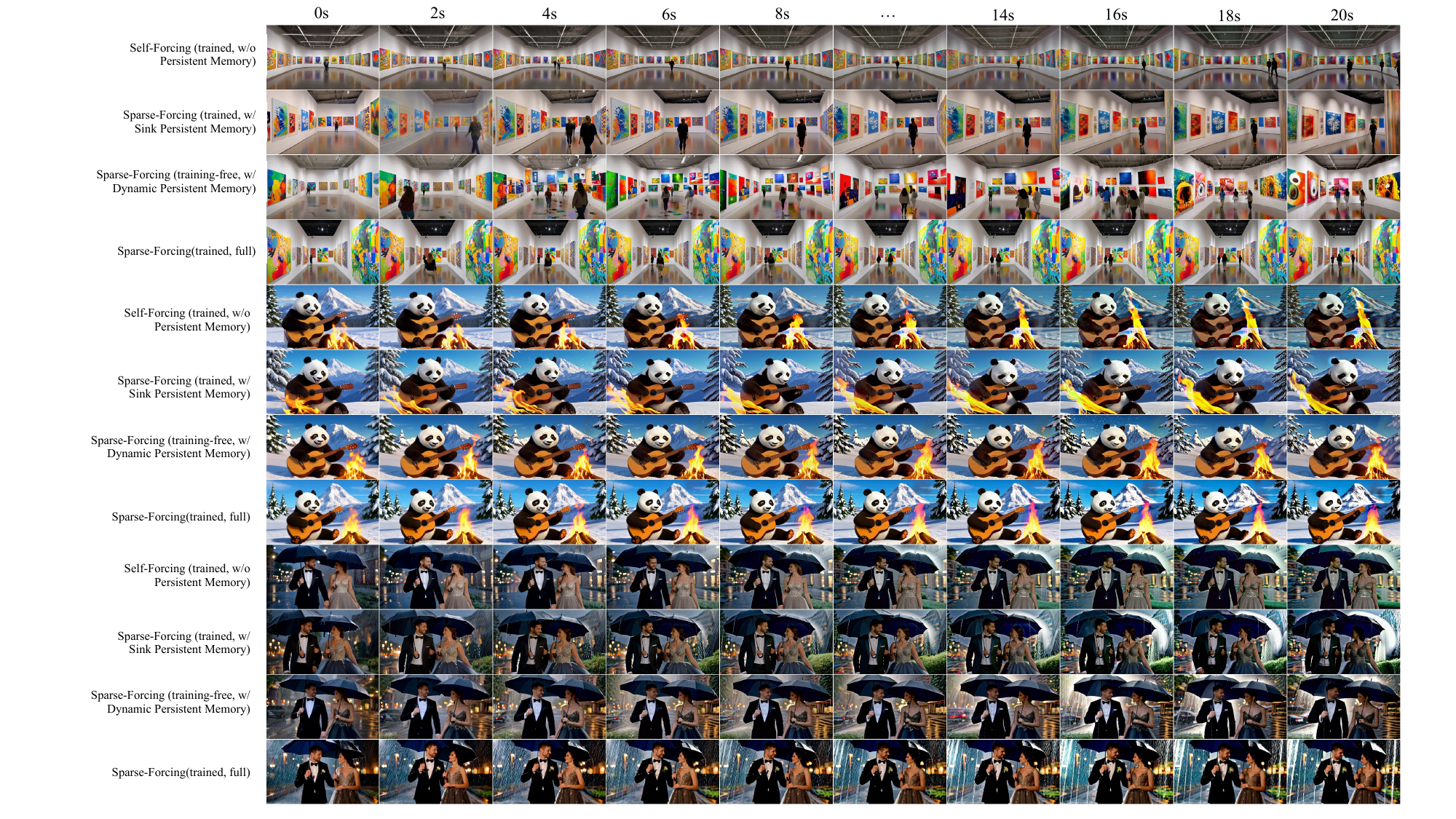}
    \caption{A comparison for different Sparse Forcing models and the baseline.}
    \label{fig:Ablation1_full}
\end{figure*}

\Cref{fig:Ablation1_full} shows a comparison for different Sparse Forcing models and the baseline.

The prompts used for generated videos are in the following:

``A modern art museum featuring a vibrant array of colorful abstract paintings. The walls are white, providing a stark contrast to the bright, expressive artworks hanging on them. Various artists' works are displayed, showcasing a mix of styles including geometric shapes, splashes of paint, and bold brushstrokes. Visitors move gracefully among the exhibits, admiring the diverse collection. The lighting is soft and diffused, enhancing the colors and textures of each piece. Wide shots capture the expansive gallery spaces, while close-ups highlight individual paintings. The atmosphere is serene and inviting, encouraging viewers to explore and appreciate the art.''

``A cheerful, fuzzy panda playing a guitar near a warm campfire. The panda has soft, black patches against a white fluffy coat, with large, expressive eyes filled with joy. It is sitting comfortably, strumming the strings with its front paws. Flames from the campfire flicker and dance, casting gentle shadows on the ground. In the background, a majestic snow-capped mountain rises, its peaks dusted with snow under a clear blue sky. The scene is captured in a medium shot, emphasizing the cozy, serene atmosphere of the winter landscape.''

``Oil painting style, depicting a couple dressed in elegant evening attire walking home under heavy rain. The man is wearing a black tuxedo with a bow tie, while the woman is in a flowing evening gown with a fitted bodice and full skirt, adorned with intricate lace and embroidery. They are holding umbrellas, but the rain is so intense that water droplets are visible around them. The background showcases a dimly lit city street with blurred lights from distant buildings. Both s are positioned close together, sharing an umbrella, with a slightly hunched posture due to the rain. The scene captures the romantic yet challenging atmosphere of a sudden downpour. Medium shot, focusing on the couple's interaction and the surrounding environment.''

\section{More Generation Samples}\label{app:more_generation_samples}

\begin{figure*}[t]  
    \centering
    \includegraphics[width=\linewidth, trim=10mm 135mm 10mm 20mm, clip]{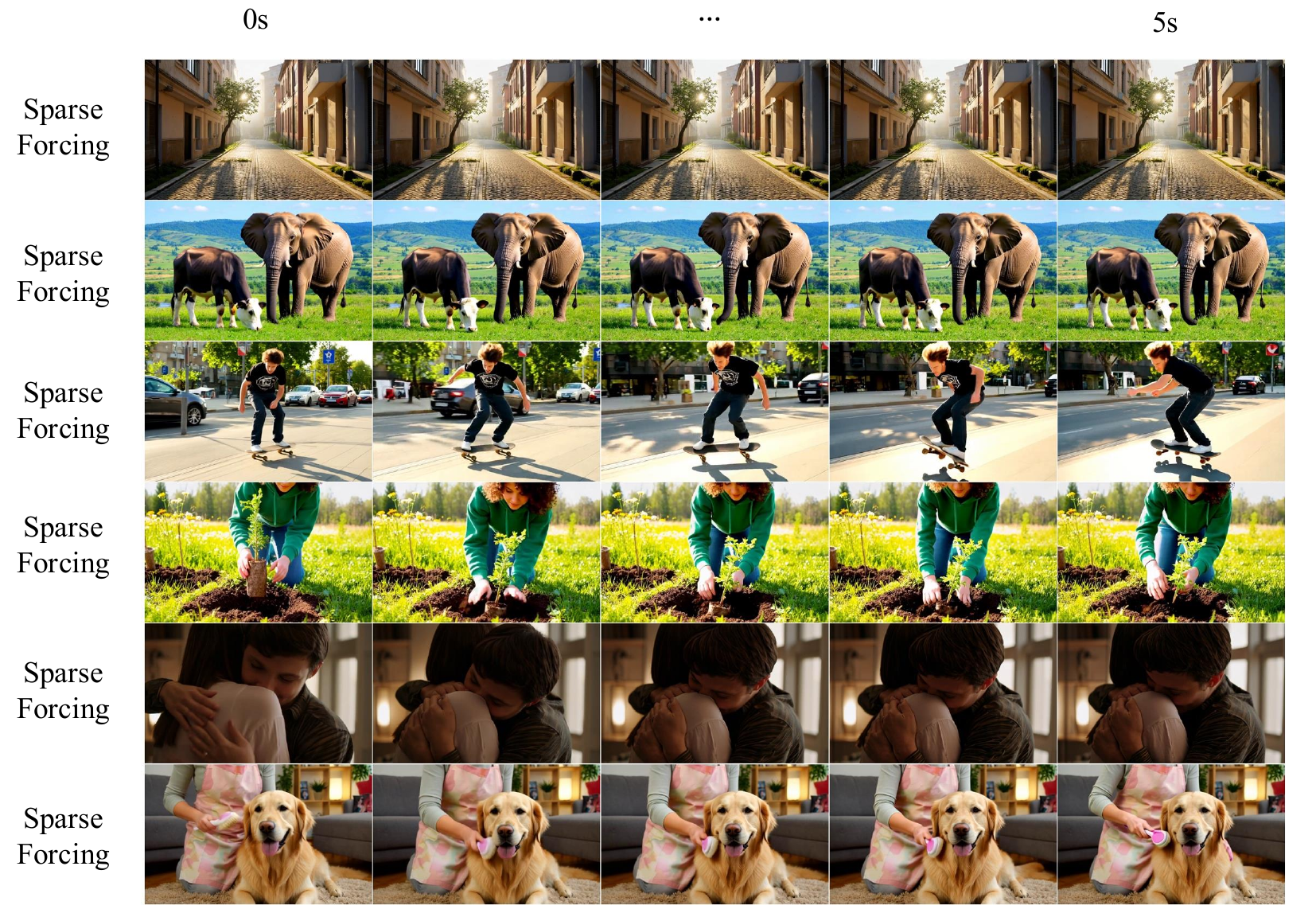}
    \caption{Samples on 5-second short-video generation for Sparse Forcing.}
    \label{fig:5s_more_examples}
\end{figure*}

We show generation samples on 5-second short video. The prompts are:

``A serene and tranquil tableau of an alley during early morning, with soft golden sunlight filtering through narrow gaps between tall buildings. The alley is clean and quiet, with cobblestone paving stones and small patches of green moss growing sporadically along the walls. A single old tree stands at one end, casting long shadows across the ground. The background showcases a mix of residential and commercial buildings, their facades weathered and painted in various pastel shades. The atmosphere is calm and peaceful, with a sense of quietude that invites contemplation. Wide shot, static scene.''

``A serene countryside landscape featuring a gentle cow grazing in the foreground and a majestic elephant standing gracefully in the background. The cow has a calm, content expression as it munches on grass, while the elephant displays a peaceful demeanor, its large ears flapping gently in the breeze. Both animals are set against a backdrop of rolling hills, lush greenery, and a clear blue sky. The cow is positioned close to the viewer, while the elephant is further away, creating depth and scale. The scene captures the natural harmony between these two distinct creatures. Medium shot focusing on both animals.''

``A young adult male is skateboarding down a city street during daytime. He has tousled brown hair, wears a black graphic t-shirt, dark blue jeans, and white sneakers. He is performing a kickflip trick, mid-air, with his skateboard rotating underneath him. The urban environment includes parked cars, street signs, and pedestrians in the background. The camera captures this action from a low angle, focusing on the skateboarder as he skillfully executes the trick. The scene is vibrant with sunlight casting shadows on the pavement.''

``A person in a green hoodie and jeans is planting trees in a sunny meadow. They are bending down to place a sapling into a freshly dug hole, then carefully covering it with soil. The person has curly hair and a determined expression. In the background, there are several other newly planted trees, and wildflowers bloom around them. The scene has a vibrant, hopeful feel, emphasizing the importance of reforestation. Medium close-up shot focusing on the person's hands and the sapling.''

``A warm and tender moment captured in a close-up shot, featuring a person embracing another person in a tight hug. Both individuals have their arms wrapped around each other, with one person's head resting gently on the other's shoulder. They appear to be sharing a loving and emotional connection. The scene is set in a cozy, dimly lit room with soft ambient lighting, creating a serene and intimate atmosphere. The focus is on the expressions of affection and comfort displayed through body language and facial expressions, conveying a sense of security and warmth.''

``A person is grooming a golden retriever in a cozy living room. The person, wearing a pastel-colored apron, gently brushes the dog's fur with a soft-bristled brush. The golden retriever is sitting obediently on a plush rug, wagging its tail occasionally. The room has warm lighting and is decorated with family photos and plants. The scene focuses on close-up shots of the person's hands working on the dog and the dog's face, showing expressions of comfort and relaxation.''

\begin{figure*}[t]
    \centering
    \includegraphics[width=0.75\linewidth, trim=10mm 18mm 10mm 20mm, clip]{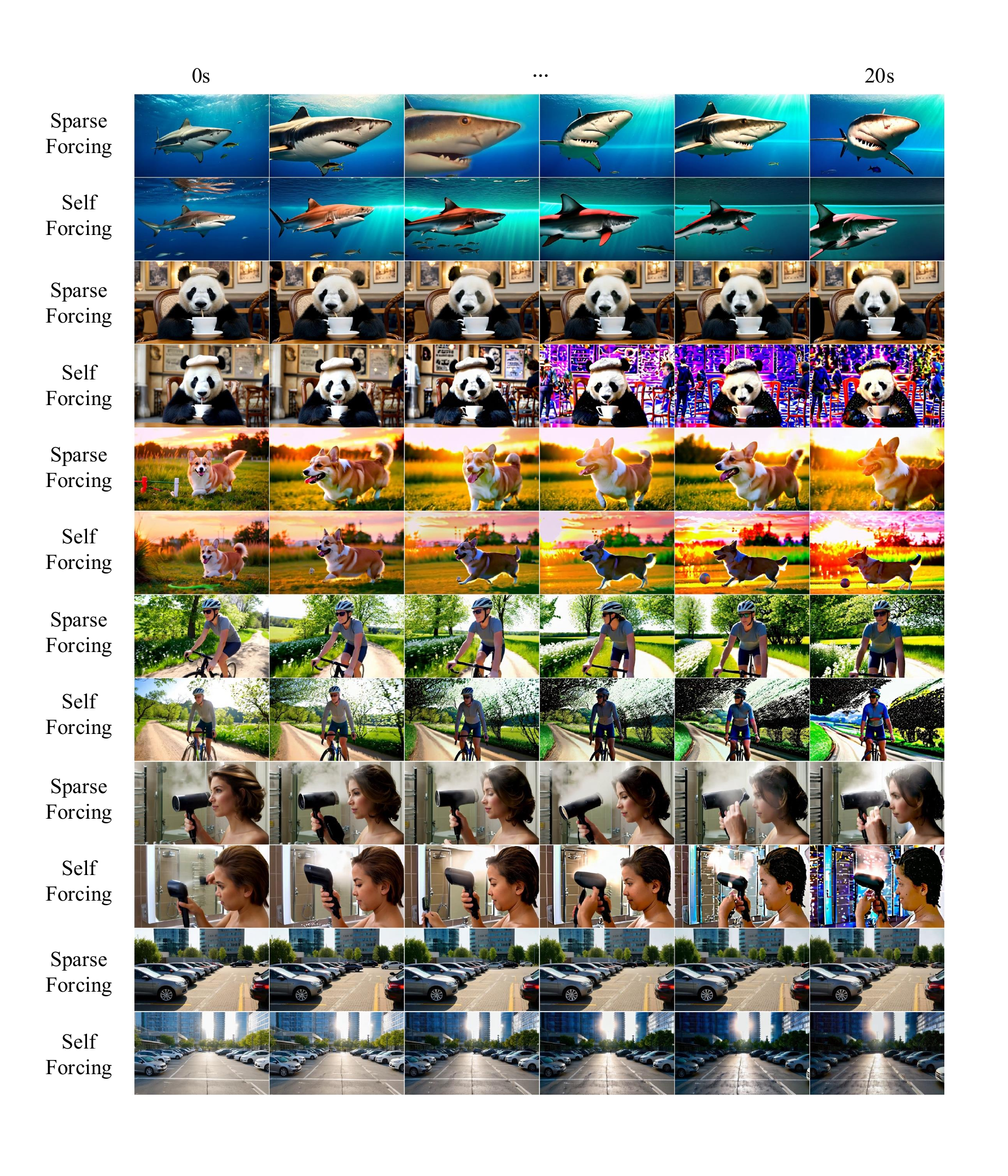}
    \caption{Samples on 20-second long-video generation for Self Forcing and Sparse Forcing.}
    \label{fig:20s_more_examples}
\end{figure*}

Additional long-video generation samples comparing Sparse Forcing with self forcing are shown in \Cref{fig:20s_more_examples}.

The prompts are:

``A large great white shark is swimming gracefully through the vast, deep blue ocean. Its sleek, muscular body cuts through the water as it propels forward with powerful tail strokes. The shark's dorsal fin slices through the surface, while smaller fish dart around it. The camera begins at a wide shot of the shark and the surrounding ocean, then smoothly zooms in to focus closely on the shark's sharp teeth and piercing eyes. The scene is filled with sunlight filtering through the water, creating a dynamic interplay of light and shadow. Close-up underwater perspective.''

``In super slow motion, a friendly panda bear sits at a cozy café table in Paris. The panda is wearing a small, stylish beret and is seated comfortably in a chair. It holds a steaming cup of coffee delicately with both paws, sipping from a straw inserted into the cup. The panda's black eyes are focused on the cup with a curious yet relaxed expression. The café background showcases elegant Parisian decor, including vintage posters and soft lighting, with other patrons subtly visible in the periphery. The scene captures the panda’s gentle movements and the delicate steam rising from the coffee in a close-up shot.''

``A joyful, playful Corgi running and frolicking in a vibrant park during sunset. The Corgi has a cheerful expression with its tail wagging excitedly as it jumps over small obstacles and chases after a ball. The dog has short legs, a sturdy build, and a fluffy coat. The background showcases a beautiful orange and pink sky with tall grass swaying gently in the breeze. The scene transitions from a wide shot of the park to a close-up of the Corgi, emphasizing its lively actions and the warm, serene atmosphere.''

``A person is cycling through a scenic park trail. The rider is wearing a helmet, casual clothes, and sunglasses, pedaling steadily. They are mid-action, leaning slightly forward, with one hand on the handlebars and the other hanging loosely. The environment around them includes lush green trees, blooming flowers, and a winding dirt path. The sun is shining brightly, casting dappled shadows through the leaves. The scene captures a close-up of the rider from a side angle, focusing on their determined expression and the motion of the bicycle wheels.''

``A close-up of a person styling their hair with a handheld hair dryer. The person, with a focused expression, holds the hair dryer in one hand and uses a brush in the other to smooth their hair. They are standing in front of a bathroom mirror, which reflects their determined face and the steam from the hair dryer. The background includes a typical bathroom setup with a towel rack and a sink. The person is mid-action, with natural motion captured in a medium shot that emphasizes the interaction between the person and the hair dryer.''

``A still frame showing a busy parking lot during a sunny day. The scene includes multiple cars of various makes and models parked neatly in rows. In the background, there are tall office buildings with glass facades reflecting sunlight. The pavement is clean and well-maintained, with clear parking lines and spaces marked. A few people can be seen walking between cars, and a couple of vehicles are driving in and out of the lot. The overall atmosphere is calm and orderly. Wide shot, static scene.''

\section{Rearrange for Locality Preserving}\label{app:locality_preserving}

\begin{algorithm}[htbp]
\caption{Locality-Preserving Spatiotemporal Block Rearrange}
\label{alg:block_reshape}
\begin{algorithmic}[1]
\REQUIRE Latent tensor $\mathbf{X}\in\mathbb{R}^{T\times H\times W\times d}$ \COMMENT{$d$: feature dim, $d=H\times d_h$}
\REQUIRE Block shape $(B_t,B_h,B_w)$
\ENSURE Block-major tensor $\mathbf{X}_{\mathrm{blk}}\in\mathbb{R}^{N_b\times B\times d}$

\STATE $N_t \leftarrow T / B_t,\;\; N_h \leftarrow H / B_h,\;\; N_w \leftarrow W / B_w$
\STATE $B \leftarrow B_t\cdot B_h\cdot B_w,\;\; N_b \leftarrow N_t\cdot N_h\cdot N_w$ \COMMENT{$B$: \#tokens per block; $N_b$: \#blocks}
\STATE $\mathbf{X} \leftarrow \mathrm{reshape}(\mathbf{X}, [N_t, B_t,\, N_h, B_h,\, N_w, B_w,\, d])$ \COMMENT{group into blocks}
\STATE $\mathbf{X} \leftarrow \mathrm{permute}(\mathbf{X}, [N_t, N_h, N_w,\, B_t, B_h, B_w,\, d])$ \COMMENT{block-contiguous layout}
\STATE $\mathbf{X}_{\mathrm{blk}} \leftarrow \mathrm{reshape}(\mathbf{X}, [N_b, B, d])$ \COMMENT{flatten to block-major}
\STATE \textbf{return} $\mathbf{X}_{\mathrm{blk}}$
\end{algorithmic}
\end{algorithm}

\Cref{alg:block_reshape} summarizes the locality-preserving reshape that converts a spatiotemporal latent tensor into a block-major layout.
Given a latent tensor $\mathbf{X}\in\mathbb{R}^{T\times H\times W\times d}$ and a block shape $(B_t, B_h, B_w)$, we partition the $T\times H\times W$ grid into $N_t=T/B_t$, $N_h=H/B_h$, and $N_w=W/B_w$ blocks, each containing $B=B_t B_h B_w$ tokens.
First, we reshape $\mathbf{X}$ into a six-dimensional view that explicitly separates block indices from intra-block coordinates, and then permutes dimensions so that block indices $(n_t,n_h,n_w)$ are contiguous.
Finally, we flatten the tensor into a matrix $\mathbf{X}_{\text{blk}}\in\mathbb{R}^{N_b\times B\times d}$ with $N_b=N_tN_hN_w$, where each row corresponds to one spatiotemporal block and tokens within a block are stored contiguously.
This layout is convenient for our blockwise sparse computation: it preserves local spatiotemporal neighborhoods, enables coalesced memory access when loading a block.
In practice, $\mathbf{X}_{\text{blk}}$ serves as the common interface between the model-side tensor representation and our block-sparse attention kernels, allowing block-level selection and compute to be implemented as contiguous reads and writes with minimal indexing overhead.

\section{Analysis and breakdown of PBSA}\label{label:PBSA_breakdown}\label{app:pbsa_latency_breakdown}

\begin{table}[tbhp]
  \centering
  \caption{Latency breakdown of \textsc{PBSA} kernel for a 65536-length KV sequence with 4096-token persistent memory and 6.25\% local block sparsity.}
  \label{tab:pbsa_latency_breakdown}
  \setlength{\tabcolsep}{8pt}
  \renewcommand{\arraystretch}{1.10}
  \begin{tabular}{l r}
    \toprule
    \textbf{Operation} & \textbf{Percentage (\%)} \\
    \midrule
    Block Compression                 & 2.51  \\
    Block Representative Attention    & 1.39  \\
    Block Representative Broadcasting & 1.73  \\
    Row-wise Top-$K$ Block Selection  & 9.52  \\
    Generate Fine-Stage Mask          & 9.52  \\
    Block Sparse Attention            & 73.16 \\
    Others                            & 2.16  \\
    \bottomrule
  \end{tabular}
\end{table}
To understand where computation is spent in the customized \textsc{PBSA} kernel, we profile the end-to-end latency and decompose it into major stages.
\Cref{tab:pbsa_latency_breakdown} reports the latency breakdown for a representative configuration with a 65,536-length KV sequence, 4,096-token persistent memory, and 6.25\% local block sparsity.
The fine-stage Block Sparse Attention dominates the runtime (73.16\%), indicating that PBSA is primarily compute-bound in the sparse attention computation rather than in selection and mask-construction overhead.
The coarse block representative pathway is lightweight: Block Representative Attention and its broadcasting together account for only 3.12\%.
In contrast, row-wise Top-$K$ block selection and fine-stage mask generation introduce a moderate overhead (19.04\%), highlighting an additional optimization opportunity on the attention kernel.

\section{Measured Latency for both forward-pass and backward-pass of PBSA on H100}\label{app:PBSA_latency}

\begin{table*}[t]
\centering
\caption{Benchmarking kernel latency for PBSA and FA2(forward and backward).}
\label{tab:kernel_benchmark_latency_fwd_bwd}
\begin{adjustbox}{max width=\textwidth}
\setlength{\tabcolsep}{3.5pt}
\renewcommand{\arraystretch}{1.15}
\begin{tabular}{@{}rrrrr|rrr|rrr@{}}
\toprule
\multirow{2}{*}{$\text{Top-}K$} &
\multirow{2}{*}{$N_C$} &
\multirow{2}{*}{$N_L/N_C$} &
\multirow{2}{*}{$N_P/N_L$} &
\multirow{2}{*}{$N_{KV}$} &
\multicolumn{3}{c|}{Forward} &
\multicolumn{3}{c}{Backward} \\
\cmidrule(lr){6-8}\cmidrule(lr){9-11}
& & & & &
PBSA(ms) & FA2(ms) & Speedup &
PBSA(ms) & FA2(ms) & Speedup \\
\midrule
0.0625 &  5376 & 2 & 0.25 &  13440 &  0.691 &  1.314 & 1.90$\times$ &  1.301 &   4.165 & 3.20$\times$ \\
0.0625 &  5376 & 2 & 0.50 &  16128 &  0.892 &  1.647 & 1.85$\times$ &  1.786 &   4.917 & 2.75$\times$ \\
0.0625 &  5376 & 4 & 0.25 &  26880 &  1.025 &  2.739 & 2.67$\times$ &  2.279 &   8.079 & 3.55$\times$ \\
0.0625 &  5376 & 4 & 0.50 &  32256 &  1.460 &  3.428 & 2.35$\times$ &  3.560 &   9.269 & 2.60$\times$ \\
0.0625 & 21504 & 2 & 0.25 &  53760 &  4.847 & 21.570 & 4.45$\times$ & 13.666 &  63.109 & 4.62$\times$ \\
0.0625 & 21504 & 2 & 0.50 &  64512 &  8.740 & 26.236 & 3.00$\times$ & 23.338 &  74.533 & 3.19$\times$ \\
0.0625 & 21504 & 4 & 0.25 & 107520 & 10.429 & 43.533 & 4.17$\times$ & 26.089 & 124.094 & 4.76$\times$ \\
0.0625 & 21504 & 4 & 0.50 & 129024 & 16.161 & 52.864 & 3.27$\times$ & 45.830 & 148.566 & 3.24$\times$ \\
\midrule
0.1250 &  5376 & 2 & 0.25 &  13440 &  0.737 &  1.368 & 1.86$\times$ &  1.431 &   4.491 & 3.14$\times$ \\
0.1250 &  5376 & 2 & 0.50 &  16128 &  0.956 &  1.658 & 1.73$\times$ &  1.985 &   4.908 & 2.47$\times$ \\
0.1250 &  5376 & 4 & 0.25 &  26880 &  1.121 &  2.731 & 2.44$\times$ &  2.544 &   8.136 & 3.20$\times$ \\
0.1250 &  5376 & 4 & 0.50 &  32256 &  1.526 &  3.529 & 2.31$\times$ &  3.871 &   9.292 & 2.40$\times$ \\
0.1250 & 21504 & 2 & 0.25 &  53760 &  5.599 & 21.642 & 3.87$\times$ & 15.670 &  63.376 & 4.04$\times$ \\
0.1250 & 21504 & 2 & 0.50 &  64512 &  9.614 & 26.170 & 2.72$\times$ & 26.071 &  75.054 & 2.88$\times$ \\
0.1250 & 21504 & 4 & 0.25 & 107520 & 12.099 & 43.353 & 3.58$\times$ & 30.399 & 124.289 & 4.09$\times$ \\
0.1250 & 21504 & 4 & 0.50 & 129024 & 18.342 & 52.885 & 2.88$\times$ & 51.079 & 148.460 & 2.91$\times$ \\
\midrule
0.2500 &  5376 & 2 & 0.25 &  13440 &  0.813 &  1.377 & 1.69$\times$ &  1.643 &   4.427 & 2.69$\times$ \\
0.2500 &  5376 & 2 & 0.50 &  16128 &  1.052 &  1.719 & 1.63$\times$ &  2.329 &   4.948 & 2.13$\times$ \\
0.2500 &  5376 & 4 & 0.25 &  26880 &  1.315 &  2.730 & 2.08$\times$ &  3.083 &   8.143 & 2.64$\times$ \\
0.2500 &  5376 & 4 & 0.50 &  32256 &  1.739 &  3.449 & 1.98$\times$ &  4.591 &   9.282 & 2.02$\times$ \\
0.2500 & 21504 & 2 & 0.25 &  53760 &  6.838 & 21.622 & 3.16$\times$ & 19.334 &  63.192 & 3.27$\times$ \\
0.2500 & 21504 & 2 & 0.50 &  64512 & 11.164 & 26.192 & 2.35$\times$ & 31.594 &  74.546 & 2.36$\times$ \\
0.2500 & 21504 & 4 & 0.25 & 107520 & 15.022 & 43.410 & 2.89$\times$ & 39.637 & 124.244 & 3.13$\times$ \\
0.2500 & 21504 & 4 & 0.50 & 129024 & 21.966 & 52.801 & 2.40$\times$ & 60.628 & 148.350 & 2.45$\times$ \\
\bottomrule
\end{tabular}
\end{adjustbox}
\end{table*}

\Cref{tab:kernel_benchmark_latency_fwd_bwd} benchmarks the averaged latency for both forward- and backward- passes in PBSA and the baseline FA2 across different configurations on Nvidia H100 96G.

\section{Broader Societal Impact}\label{app:broader_impact}
This work improves the efficiency, scalability and generation quality of both short-horizon and long-horizon video generation by introducing Sparse Forcing and an optimized sparse-attention kernel. By reducing peak KV-cache usage and accelerating decoding stages, it can lower the compute barrier for video generation, enabling broader access for research and creative applications and potentially reducing energy consumed per generated samples when replacing more expensive inference.


\section{Reproducibility and Limitations}
We will release code, training and evaluation recipes, and our PBSA kernel implementation to facilitate reproducibility and adoption. A limitation of the current work is that we evaluate Sparse Forcing on a single pretrained video diffusion backbone and a fixed resolution; extending our analysis to other backbones and higher resolutions is an interesting direction for future work.

\end{document}